\documentclass{article}

\usepackage{microtype}
\usepackage{graphicx}
\usepackage{subcaption}
\usepackage{booktabs} 
\usepackage{hyperref}

\usepackage[preprint]{icml2026}

\usepackage{amsmath}
\usepackage{amssymb}
\usepackage{mathtools}
\usepackage{amsthm}
\usepackage{multirow}

\usepackage[capitalize,noabbrev]{cleveref}

\newcommand{\MetaTTRL}{\textbf{Meta-TTRL}} 
\usepackage{graphicx} 

\usepackage{xcolor}      
\usepackage{colortbl}     
\usepackage{hhline}   
\usepackage{makecell}   
\usepackage{textcomp}
\arrayrulecolor{black}
\definecolor{AliceBlue}{RGB}{240, 248, 255}

\definecolor{lightgray}{gray}{0.95}
\definecolor{deepblue}{RGB}{70,130,180}
\definecolor{deepgray}{RGB}{119,136,153}
\definecolor{RosyBrown}{RGB}{188,143,143}
\definecolor{PeachPuff3}{RGB}{205,175,149}
\definecolor{ForestGreen}{RGB}{0, 128, 0}
\definecolor{linkblue}{RGB}{0, 30, 110}

\theoremstyle{plain}

\theoremstyle{definition}

\theoremstyle{remark}

\usepackage[textsize=tiny]{todonotes}

\icmltitlerunning{Meta-TTRL}

\begin{document}

\twocolumn[
  \icmltitle{Meta-TTRL: A Metacognitive Framework for Self-Improving Test-Time \\ Reinforcement Learning in Unified Multimodal Models}
  \icmlsetsymbol{equal}{*}
  \begin{icmlauthorlist}
    \icmlauthor{Lit Sin Tan}{equal,school,comp}
    \icmlauthor{Junzhe Chen}{equal,school}
    \icmlauthor{Xiaolong Fu}{comp}
    \icmlauthor{Lichen Ma}{comp}
    \icmlauthor{Junshi Huang}{comp}
    \icmlauthor{Jianzhong Shi}{comp}
    \icmlauthor{Yan Li}{comp}
    \icmlauthor{Lijie Wen}{school}
  \end{icmlauthorlist}
  
  \icmlaffiliation{school}{Tsinghua University}
  \icmlaffiliation{comp}{JD.COM}
  \icmlcorrespondingauthor{Lijie Wen}{wenlj@tsinghua.edu.cn}
  
  \icmlkeywords{Machine Learning, ICML}
  \vskip 0.3in
]
\printAffiliationsAndNotice{\icmlEqualContribution}

\begin{abstract}
    Existing test-time scaling (TTS) methods for unified multimodal models (UMMs) in text-to-image (T2I) generation primarily rely on search or sampling strategies that produce only instance-level improvements, limiting the ability to learn from prior inferences and accumulate knowledge across similar prompts. To overcome these limitations, we propose \textbf{Meta-TTRL}, a metacognitive test-time reinforcement learning framework. \textbf{Meta-TTRL} performs test-time parameter optimization guided by 
    model-intrinsic monitoring signals derived from the meta-knowledge of UMMs, achieving \emph{self-improvement} and capability-level improvement at test time. Extensive experiments demonstrate that \textbf{Meta-TTRL} generalizes well across three representative UMMs, including Janus-Pro-7B, BAGEL, and Qwen-Image, achieving significant gains on compositional reasoning tasks and multiple T2I benchmarks with limited data. We provide the first comprehensive analysis to investigate the potential of test-time reinforcement learning (TTRL) for T2I generation in UMMs. Our analysis further reveals a key insight underlying effective TTRL: \emph{metacognitive synergy}, where monitoring signals align with the model’s optimization regime to enable self-improvement.
\end{abstract}

\section{Introduction}
Unified multimodal models (UMMs) have recently emerged as a powerful paradigm for text-to-image (T2I) generation by integrating multimodal understanding and generation within a single architecture~\citep{xie2024show, chen2025januspro, deng2025emerging, wu2025qwenimagetechnicalreport, wang2024emu3, cui2025emu3}. This unified architecture enables coherent semantic grounding and flexible visual synthesis, leading to strong performance in both image quality and compositional fidelity.
\begin{figure}[t]
    \centering
    \includegraphics[width=0.999\linewidth]{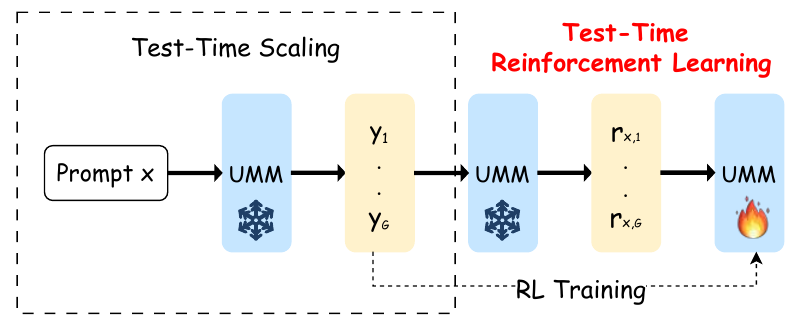} 
    \caption{Test-Time Reinforcement Learning in UMMs.}
    \label{fig:TTRL_position} 
    \vspace{-10pt} 
\end{figure}

 The rapid progress of UMMs has been largely driven by the success of scaling large pre-trained language models (LLMs)~\citep{brown2020language, achiam2023gpt}. The strong representational and reasoning capabilities enabled by LLM scaling have inspired the adoption of similar scaling strategies in UMMs~\citep{chen2025januspro, cui2025emu3}. However, this scaling-centric paradigm has a significant downside: the computational cost of training and deploying ever-larger UMMs becomes prohibitively expensive, limiting their practical scalability and accessibility.

 To mitigate the computational cost of model scaling, recent work has focused on test-time scaling (TTS) methods, which aim to enhance model performance by allocating additional computational resources during test time. Recent studies have shown that TTS can improve T2I generation performance in UMMs~\citep{tian2025unigen, guo2025can, liangyu2026unit, wang2025imagent}. TTS approaches for T2I generation in UMMs can be generally categorized into two paradigms. (1) Parallel sampling-based scaling~\citep{tian2025unigen, guo2025can}, where multiple candidate images are generated through sampling and the best one is selected via Best-of-$N$ selection based on verification scores. Candidate ranking in this paradigm generally relies on two types of verification scores: \emph{external reward models}~\citep{guo2025can}, which employ outcome reward models or process reward models to assess the alignment between the generated image and the prompt; and \emph{self-verification}~\citep{tian2025unigen}, in which the UMM is specifically trained to serve as the verifier, thereby acquiring chain-of-thought-based verification capability. (2) Sequential iterative scaling~\citep{liangyu2026unit, wang2025imagent}, where the UMM progressively refines outputs across multiple inference rounds through reasoning, evaluation, and refinement steps, enabling iterative improvement based on intermediate results.
 
However, current TTS approaches for T2I generation in UMMs improve generation performance primarily through test-time sampling or search strategies while keeping the model parameters fixed. As a result, the additional computation yields only \emph{instance-level improvements}, and the UMM fails to accumulate knowledge from prior inference experiences. In other words, any performance gain is \emph{ephemeral}: when encountering similar prompts in subsequent queries, the UMM often exhibits the same failure modes and must repeat the entire TTS procedure again. This observation naturally raises a key question: \emph{Can the experience obtained during test-time scaling be leveraged for learning?}

A promising direction is test-time reinforcement learning (TTRL), which leverages test-time experience as learning signals to update model parameters at test time. TTRL has recently shown strong potential in language reasoning tasks~\citep{zuo2025ttrl} and vision--language understanding tasks~\citep{singh2025ttrv}. However, despite its success in these tasks, TTRL for T2I generation in UMMs remains largely unexplored. This leads to a key question:
\emph{Can reinforcement learning enable effective and robust test-time learning for T2I generation in UMMs without relying on external reward models?}

To answer this question, we propose \textbf{Meta-TTRL}, a metacognitive test-time reinforcement learning framework for T2I generation in UMMs. As illustrated in Figure~\ref{fig:TTRL_position}, \textbf{Meta-TTRL} extends TTS by leveraging the experience obtained during TTS as learning signals to update model parameters via reinforcement learning. The framework introduces a two-level metacognitive architecture in which the UMM operates at two levels: an object-level generator produces candidate images, while a meta-level introspector constructs structured evaluation rubrics and monitors generation outputs. The model-intrinsic monitoring signals, derived from the meta-knowledge of UMMs, eliminate the need for external reward models while providing robust supervision and enabling self-improvement. Through the interaction between meta-level monitoring and object-level policy optimization, \textbf{Meta-TTRL} forms a monitoring–control loop that enables capability-level improvement, allowing performance gains to persist when the model encounters similar prompts.

Beyond proposing a new framework, we aim to better understand the potential of TTRL for T2I generation in UMMs. Through comprehensive analysis, we identify a key insight underlying effective TTRL, referred to as \emph{metacognitive synergy}. Specifically, robust and effective TTRL emerges when monitoring signals are well aligned with the model’s optimization regime and model’s capacity for self-improvement. This finding suggests that the effectiveness of TTRL is determined less by the absolute capability or scale of introspectors, and more by how well monitoring signals align with the model’s optimization regime.

Our contributions can be summarized as:
\begin{itemize}

\item \textbf{Meta-TTRL Framework.}
We propose \textbf{Meta-TTRL}, the first test-time reinforcement learning framework for T2I generation in UMMs. \textbf{Meta-TTRL} leverages model-intrinsic monitoring signals derived from the meta-knowledge of UMMs to guide test-time policy optimization, enabling self-improvement.

\item \textbf{Comprehensive Analysis of TTRL.}
We conduct analyses to study the potential of TTRL from three aspects: \emph{external introspector (E-TTRL)}, \emph{RL leakage}, and \emph{alternative monitoring signals}. These analyses reveal a key insight underlying effective TTRL, termed \emph{metacognitive synergy}: robust and effective TTRL emerges when monitoring signals are well aligned with the model’s optimization regime.

\item \textbf{Strong Empirical Improvements.}
Extensive experiments show that \textbf{Meta-TTRL} consistently improves T2I generation performance across three representative UMMs: Janus-Pro-7B, BAGEL, and Qwen-Image. \textbf{Meta-TTRL} yields significant improvements in compositional reasoning tasks (e.g., shape, texture, and spatial reasoning), with relative gains exceeding 50\% for weaker base models while also providing consistent benefits for stronger models. In addition, the improvements generalize well to out-of-distribution settings, including cross-benchmark evaluations, indicating that \textbf{Meta-TTRL} does not overfit to specific benchmarks.
\end{itemize}

\section{Related Work}
\paragraph{Unified Multimodal Models (UMMs).} 
Unified multimodal models (UMMs) jointly support multimodal understanding and generation. Autoregressive methods extend the next-token prediction paradigm to jointly model text and discrete visual tokens~\citep{team2024chameleon, wang2024emu3, chen2025januspro}. Diffusion methods couple language models with diffusion modules for image generation~\citep{dong2023dreamllm, wu2025qwenimagetechnicalreport}. Unified transformer architectures integrate language modeling and diffusion within single architectures~\citep{yu2024mmoe, deng2025emerging}. Our framework generalizes across all three architectural paradigms and is evaluated on three representative UMMs, including Janus-Pro-7B, Qwen-Image, and BAGEL.

\begin{figure*}[h]
    \centering
    \includegraphics[width=\textwidth]{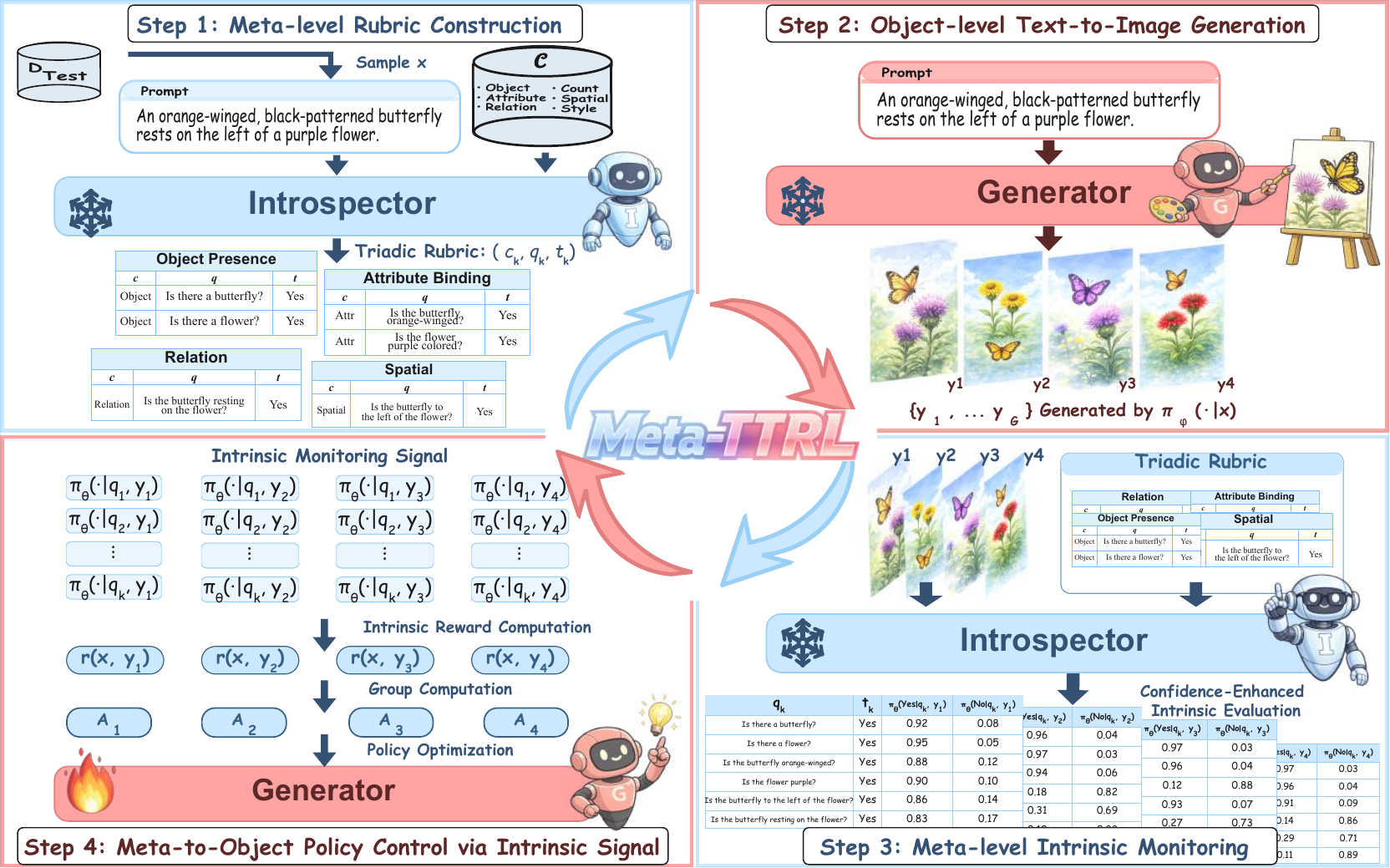} \caption{\MetaTTRL\ operates with a \textbf{meta-level} introspector that constructs rubric and provides intrinsic monitoring signals, and an \textbf{object-level} generator that produces candidate images.
These intrinsic signals are aggregated as rewards to guide policy optimization and improve T2I generation. For clarity, the figure omits the question index $m$ and shows only representative verification questions.
}
    \label{fig:method} 
    \vspace{-10pt} 
\end{figure*}

\paragraph{Test-Time Reinforcement Learning (TTRL).}
Test-time reinforcement learning (TTRL) generalizes test-time training by leveraging reinforcement learning with weak proxy rewards~\citep{zuo2025ttrl, singh2025ttrv}. In language reasoning, TTRL with majority voting can approach the performance of models trained on labeled test data~\citep{zuo2025ttrl}. Despite these advances, applying TTRL to T2I generation remains largely underexplored. In this work, we introduce \textbf{Meta-TTRL}, a metacognitive TTRL framework designed for T2I generation in UMMs.

\section{Methodology}
\label{sec:method}
In this section, we present \textbf{Meta-TTRL}. A core design principle of \textbf{Meta-TTRL} is to leverage the meta-knowledge of UMMs. At test time, the UMM performs intrinsic metacognitive monitoring based on meta-knowledge, evaluating its own outputs according to its internal multimodal reasoning and understanding. The resulting monitoring signals are then used to guide test-time policy optimization. Following the Nelson--Narens model of metacognition~\cite{nelson1990metamemory}, \textbf{Meta-TTRL} is formulated as a metacognitive framework built upon a two-level architecture. 

\subsection{Two-level Metacognitive Architecture}
\label{subsec:architecture}
\textbf{Meta-TTRL} implements a two-level architecture consisting of an \textbf{object level} and a \textbf{meta level}.
The \textbf{object level} performs the core text-to-image generation task, while the \textbf{meta level} monitors the outputs of the object level and produces intrinsic monitoring signals. Formally, let $\mathcal{X}$ denote the space of prompts and $\mathcal{Y}$ the space of images.
Given a test prompt $x \in \mathcal{X}$, the UMM operates as follows. At the object level, the UMM functions as a \emph{generator} parameterized by a policy $\pi_\phi(\cdot \mid x)$, which produces candidate images $y \in \mathcal{Y}$. At the meta level, the UMM functions as an \emph{introspector} parameterized by $\theta$, which evaluates prompt–image pairs $(x,y)$ and produces intrinsic monitoring signals used for policy optimization. This separation provides the structural basis for the monitoring–control loop described next.

\subsection{Monitoring--Control Loop}
\label{subsec:monitoring_control}
Building on the two-level architecture, \textbf{Meta-TTRL} operates through a closed-loop interaction between monitoring and control. At each iteration, the object-level \emph{generator} samples candidate images $y \sim \pi_\phi(\cdot \mid x).$ The meta-level \emph{introspector} then evaluates each prompt–image pair $(x,y)$ and maps it to a structured evaluation space, producing an intrinsic reward $r(x,y)$,  which serves as the monitoring signal. Based on this monitoring signal, the meta level controls the \emph{generator} by adjusting the \emph{generator} policy to maximize the reward:
\begin{equation}
\max_{\phi} \; \mathbb{E}_{y \sim \pi_\phi(\cdot \mid x)} [r(x, y)].
\label{eq:objective}
\end{equation}
In practice, the \emph{generator} policy is updated via gradient-based optimization:
\begin{equation}
\phi \leftarrow \phi + \eta \nabla_\phi \mathbb{E}_{y \sim \pi_\phi(\cdot \mid x)} [r(x, y)],
\label{eq:update}
\end{equation}
where $\eta$ denotes the learning rate. This monitoring–control loop enables the UMM to iteratively refine its generation policy through intrinsic monitoring at test time. Building on this formulation, we implement the monitoring–control loop as a four-step procedure illustrated in Figure~\ref{fig:method}.

\begin{table*}[ht]
\small
\centering
\caption{\textbf{Evaluation results on TIIF-Bench \cite{wei2025tiif}, T2I-CompBench++ \cite{huang2025t2i}, DPG-Bench \cite{hu2024ella}}. Arrows ($\uparrow$) denote that higher is better. The T2I-CompBench++ score is normalized to a 0–100 scale for visualization.}
\renewcommand{\arraystretch}{1.5}
\setlength{\tabcolsep}{2.1pt}      
\definecolor{mygreen}{HTML}{EDFCED}
\definecolor{tabgray}{RGB}{245,245,245}

\resizebox{\textwidth}{!}{

\begin{tabular*}{\textwidth}{@{\extracolsep{\fill}}lcc|cccccccc|c@{}}

\Xhline{1.5pt}
\multirow{2}{*}{Model}
& \multicolumn{2}{c|}{\textbf{TIIF-Bench} $\uparrow$}
& \multicolumn{8}{c|}{\textbf{T2I-CompBench++} $\uparrow$}  
& \textbf{DPG} $\uparrow$  \\
\cline{2-12}
& Short & Long 
& \makecell{Color}
& \makecell{Shape}
& \makecell{Texture}
& \makecell{2D\\Spatial}
& \makecell{3D\\Spatial}
& \makecell{Non-\\Spatial}
& \makecell{Numeracy}
& \makecell{Complex}

& \textbf{Score}  \\
\hline

Qwen-Image  

&83.45 &82.48 

&82.80 &59.86 &74.51 
&44.30 &45.36 
&31.71 
&75.82 
&38.38 

&88.32

\\

\textbf{w/ Meta-TTRL}    
&85.28 &84.67

&84.50 &62.42 &76.36 
&46.59 &46.11 
&31.85 
&76.77 
&38.48 

&89.00

\\
\rowcolor{tabgray}
\textbf{$\triangle$} &\textbf{\textcolor{ForestGreen}{+1.83}} & \textbf{\textcolor{ForestGreen}{+2.19}} & \textbf{\textcolor{ForestGreen}{+1.70}} & \textbf{\textcolor{ForestGreen}{+2.56}} & \textbf{\textcolor{ForestGreen}{+1.85}} & \textbf{\textcolor{ForestGreen}{+2.29}} & \textbf{\textcolor{ForestGreen}{+0.75}} & \textbf{\textcolor{ForestGreen}{+0.14}} & \textbf{\textcolor{ForestGreen}{+0.95}} & \textbf{\textcolor{ForestGreen}{+0.10}} & \textbf{\textcolor{ForestGreen}{+0.68}}  
\\
\rowcolor{tabgray}
& \small $ \uparrow 2.19\% $ & \small $ \uparrow 2.66\% $ 
& \small $ \uparrow 2.05\% $ & \small $ \uparrow 4.28\% $ & \small $ \uparrow 2.48\% $ 
& \small $ \uparrow 5.17\% $ & \small $ \uparrow 1.65\% $ 
& \small $ \uparrow 0.44\% $ 
& \small $ \uparrow 1.25\% $ 
& \small $ \uparrow 0.26\% $ 
& \small $ \uparrow 0.77\% $

\\

\hline
BAGEL             
&71.65 &70.32

&80.81 &57.94 &71.54 
&34.03 &37.78 
&31.11 
&65.52 
&37.78 

&84.03 

\\
\textbf{w/ Meta-TTRL}    
&75.98 &74.59 

&82.69 &61.94 &74.55 
&38.08 &43.69 
&31.28 
&67.00 
&38.79 

&86.33

\\
\rowcolor{tabgray}
\textbf{$\triangle$} 
&\textbf{\textcolor{ForestGreen}{+4.33}} & \textbf{\textcolor{ForestGreen}{+4.27}} 

& \textbf{\textcolor{ForestGreen}{+1.88}} & \textbf{\textcolor{ForestGreen}{+4.00}} & \textbf{\textcolor{ForestGreen}{+3.01}} & \textbf{\textcolor{ForestGreen}{+4.05}} & \textbf{\textcolor{ForestGreen}{+5.91}} & \textbf{\textcolor{ForestGreen}{+0.17}} & \textbf{\textcolor{ForestGreen}{+1.48}} & \textbf{\textcolor{ForestGreen}{+1.01}} & \textbf{\textcolor{ForestGreen}{+2.30}} \\

\rowcolor{tabgray}
& $ \uparrow 6.04\% $ & $ \uparrow 6.07\% $ 
& $ \uparrow 2.33\% $ & $ \uparrow 6.90\% $ & $ \uparrow 4.21\% $ 
& $ \uparrow 11.90\% $ & $ \uparrow 15.64\% $ 
& $ \uparrow 0.55\% $ 
& $ \uparrow 2.26\% $ 
& $ \uparrow 2.67\% $ 
& $ \uparrow 2.74\% $
\\

\hline
Janus-Pro-7B   
&64.41 &66.94 

&52.02 &32.72 &40.39 
&15.73 &27.37 
&31.27 
&44.50 
&37.98 

& 83.84

\\

\textbf{w/ Meta-TTRL}   
&71.42 &70.74 

&79.33 &50.10 &67.52 
&32.46 &37.51 
&31.35 
&54.08 
&38.88 

&85.67

\\
\rowcolor{tabgray}
\textbf{$\triangle$} 
&\textbf{\textcolor{ForestGreen}{+7.01}} & \textbf{\textcolor{ForestGreen}{+3.80}} 

& \textbf{\textcolor{ForestGreen}{+27.31}} & \textbf{\textcolor{ForestGreen}{+17.38}} & \textbf{\textcolor{ForestGreen}{+27.13}} & \textbf{\textcolor{ForestGreen}{+16.73}} & \textbf{\textcolor{ForestGreen}{+10.14}} & \textbf{\textcolor{ForestGreen}{+0.08}} & \textbf{\textcolor{ForestGreen}{+9.58}}
& \textbf{\textcolor{ForestGreen}{+0.90}}  
& \textbf{\textcolor{ForestGreen}{+1.83}}
\\
\rowcolor{tabgray}
\newcommand{\improvement}[1]{\texttt{\uparrow} #1\%}
& $ \uparrow 10.88\% $ & $ \uparrow 5.68\% $ 
& $ \uparrow 52.50\% $ & $ \uparrow 53.12\% $ & $ \uparrow 67.17\% $ 
& $ \uparrow 106.36\% $ & $ \uparrow 37.04\% $ 
& $ \uparrow 0.26\% $ 
& $ \uparrow 21.53\% $ 
& $ \uparrow 2.37\% $ 
& $ \uparrow 2.18\% $

\\

\Xhline{1pt}
\end{tabular*}

}
\vspace{-10pt} 
\label{tab: main_result}
\end{table*}

\subsection{Step 1: Meta-level Rubric Construction}
Direct evaluation of complex prompts often leads to ambiguous or unstable reward signals~\cite{mehrabi2023resolving, ba2025enhancing}. \textbf{Meta-TTRL} therefore introduces a rubric construction stage that decomposes prompts into structured evaluation criteria. Given the input prompt $x$, the meta-level \emph{introspector} constructs an introspective rubric $\mathcal{Q}(x)$. The rubric $\mathcal{Q}(x)$ is constructed using a rubric schema that defines a set of cognitive dimensions:
\[
\mathcal{C}=\{\text{Object},\ \text{Attribute},\ \text{Count},\ \text{Spatial},\ \text{Relation},\ \text{Style}\}.
\]
For each dimension $c_k \in \mathcal{C}$ that is applicable to the prompt $x$, the \emph{introspector} derives a set of binary verification questions $q_{k,m}$ with corresponding target answers $t_{k,m} \in \{\text{Yes}, \text{No}\}$.
This results in a structured rubric:
\begin{equation}
\mathcal{Q}(x) = \Bigl\{ \bigl( c_k,\ \{ q_{k,m},\, t_{k,m} \}_{m=1}^{M_k} \bigr) \Bigr\}_{k=1}^K,
\end{equation}
where $M_k$ denotes the number of verification questions for dimension $c_k$. The rubric $\mathcal{Q}(x)$ defines a set of semantic constraints for intrinsic evaluation, which provide structured criteria for computing the intrinsic reward $r(x,y)$ in Eq.~\ref{eq:objective}. Through rubric construction, the model’s meta-knowledge makes implicit generation intent explicit as evaluation criteria that serve as verifiable conditions for intrinsic monitoring (see Appendix \textcolor{linkblue}{A} for rubric construction details).

\subsection{Step 2: Object-level Text-to-Image Generation}
Given the prompt $x$, the object-level \emph{generator} samples a group of candidate images from the generation policy:
\begin{equation}
\{y_1, y_2, \dots, y_G\} \sim \pi_\phi(\cdot \mid x).
\end{equation}
These samples represent diverse generations under the current policy and form the candidate outputs that will be evaluated using the rubric $\mathcal{Q}(x)$ constructed in Step~1.

\subsection{Step 3: Meta-level Intrinsic Monitoring}
\label{sec:step3}
Using the rubric $\mathcal{Q}(x)$ constructed in Step~1, the \emph{introspector} evaluates each candidate image $y_i$ generated in Step~2. For each candidate image $y_i$ and each rubric criterion $(c_k, q_{k,m}, t_{k,m})$, the \emph{introspector} estimates the confidence that the image satisfies the corresponding semantic constraint $q_{k,m}$. Specifically, a confidence-enhanced score $s_{k,m}(y_i)$ is computed as:
\begin{equation}
s_{k,m}(y_i) = \frac{\pi_\theta(t_{k,m} \mid q_{k,m}, y_i)}{\sum_{t' \in \{\text{Yes}, \text{No}\}} \pi_\theta(t' \mid q_{k,m}, y_i)},
\end{equation}
where $\pi_\theta$ denotes the \emph{introspector} policy. The intrinsic reward is then defined as the aggregated score across all rubric criteria:
\begin{equation}
r(x, y_i) = \exp \left( \frac{1}{\sum_{k=1}^K M_k} \sum_{k=1}^K \sum_{m=1}^{M_k} \log s_{k,m}(y_i) \right),
\end{equation}
This reward is obtained by using the \emph{introspector}’s meta-knowledge to evaluate how well the generated image satisfies the rubric criteria, thereby providing a monitoring signal for the subsequent meta-level control stage.

\subsection{Step 4: Meta-to-Object Policy Control via Intrinsic Signal}
Using the intrinsic reward $r(x, y_i)$ computed in Step~3, the \emph{generator} policy $\pi_\phi$ is updated using meta-level intrinsic feedback to guide object-level generation. Direct optimization of absolute reward values can be unstable due to potential bias in self-evaluation. To improve robustness, we adopt Group Relative Policy Optimization (GRPO), which updates the policy based on the relative performance of multiple candidate generations. The policy update objective is defined as:
\begin{equation}
\mathcal{J}(\phi) = \frac{1}{G} \sum_{i=1}^G \left[ \rho_i(\phi) A_i - \beta D_{\mathrm{KL}}(\pi_\phi \| \pi_{\mathrm{ref}}) \right],
\label{eq:grpo}
\end{equation}
where the relative advantage $A_i$ for image $y_i$ is defined as:
\begin{equation}
A_i =
\frac{r(x, y_i) - \mathrm{mean}(\{r(x, y_j)\}_{j=1}^G)}
{\mathrm{std}(\{r(x, y_j)\}_{j=1}^G)},
\end{equation}
and $\rho_i(\phi)$ denotes the importance sampling ratio between the current policy and the previous policy for sample $y_i$. $D_{\mathrm{KL}}$ is a KL divergence term that regularizes the policy by constraining it to remain close to the reference policy $\pi_{\mathrm{ref}}$. Intrinsic feedback derived from the \emph{introspector}’s understanding and reasoning, grounded in meta-knowledge, is used to guide the \emph{generator}’s policy updates. This allows meta-knowledge to shape object-level generation through rubric construction and intrinsic evaluation, enabling test-time self-improvement without external supervision.

\begin{algorithm}[H]
  \caption{Meta-TTRL Framework}
  \label{alg:metattrl}
  \begin{algorithmic}
    \STATE {\bfseries Input:} Test prompt $x \in \mathcal{X}$
    \STATE {\bfseries Initialize:} Introspector $\theta \leftarrow \phi_0$ (frozen initialization), generator $\pi_\phi$, rubric schema, maximum steps $T$
    \STATE Initialize step counter $t = 0$
    
    \WHILE{$t < T$ and reward has not converged}
      
      \STATE \textbf{Step 1: Meta-level Rubric Construction.} 
      
      \STATE  Construct rubric $\mathcal{Q}(x)$ from the rubric schema for $x$.
      
      \STATE \textbf{Step 2: Object-level Text-to-Image Generation.} 
      
      \STATE Sample candidates $\{y_1, y_2, \dots, y_G\} \sim \pi_\phi(\cdot \mid x)$
      
      \STATE \textbf{Step 3: Meta-level Intrinsic Monitoring.} 
      
      \STATE Evaluate candidates by introspector $\theta$:
      \FOR{each candidate $y_i$}
        \STATE Compute score $s(y_i)$ for each rubric criterion
        \STATE Aggregate scores to obtain intrinsic reward $r(x, y_i)$
      \ENDFOR
      
      \STATE \textbf{Step 4: Meta-to-Object Policy Control via Intrinsic Signal.}
      
      \STATE Control the generator via GRPO:
      \STATE Compute relative advantage $A_i$
      \STATE Form the GRPO objective ${\mathcal{J}}(\phi)$ in Eq.~\ref{eq:grpo} 
      \STATE Update $\phi \leftarrow \phi + \eta \nabla_\phi \mathcal{J}(\phi)$

      \STATE Increment step counter: $t \leftarrow t + 1$
    \ENDWHILE
  \end{algorithmic}
\end{algorithm}
\vspace{-20pt}

\section{Experiments}
\subsection{Experimental Setup}
\textbf{Models.} To evaluate the generalizability of \textbf{Meta-TTRL} across different architectural paradigms and model scales, we conduct experiments on three representative UMMs: Janus-Pro-7B~\cite{chen2025januspro}, BAGEL~\cite{deng2025emerging}, and Qwen-Image~\cite{wu2025qwenimagetechnicalreport}. Our objective is to demonstrate that \textbf{Meta-TTRL} consistently improves T2I performance across diverse UMM architectures, including costly post-trained UMMs.

\textbf{Benchmarks.} We evaluate \textbf{Meta-TTRL} on three challenging T2I benchmarks: TIIF-Bench \cite{wei2025tiif}, T2I-CompBench++ \cite{huang2025t2i}, and DPG-Bench \cite{hu2024ella}. 

\textbf{Baselines.} Since TTRL has not been previously explored for T2I generation, we primarily compare \textbf{Meta-TTRL} with the corresponding base models to evaluate whether it can achieve effective improvements through self-improvement. Additional comparisons with other models are provided in Appendix \textcolor{linkblue}{B}.

\textbf{Implementation Details.} 
All implementation details, including test-time update rules and GRPO configurations for each base models, are provided in Appendix \textcolor{linkblue}{C}.

\subsection{Main Results}
Table~\ref{tab: main_result} reports the results of \textbf{Meta-TTRL} on three challenging T2I benchmarks. In addition, figure~\ref{fig:case_study} shows qualitative case studies highlighting improvements in compositional fidelity and spatial reasoning. From these comparisons, we derive the following key conclusions:

\begin{figure}[h]
    \centering
    \includegraphics[width=\linewidth]{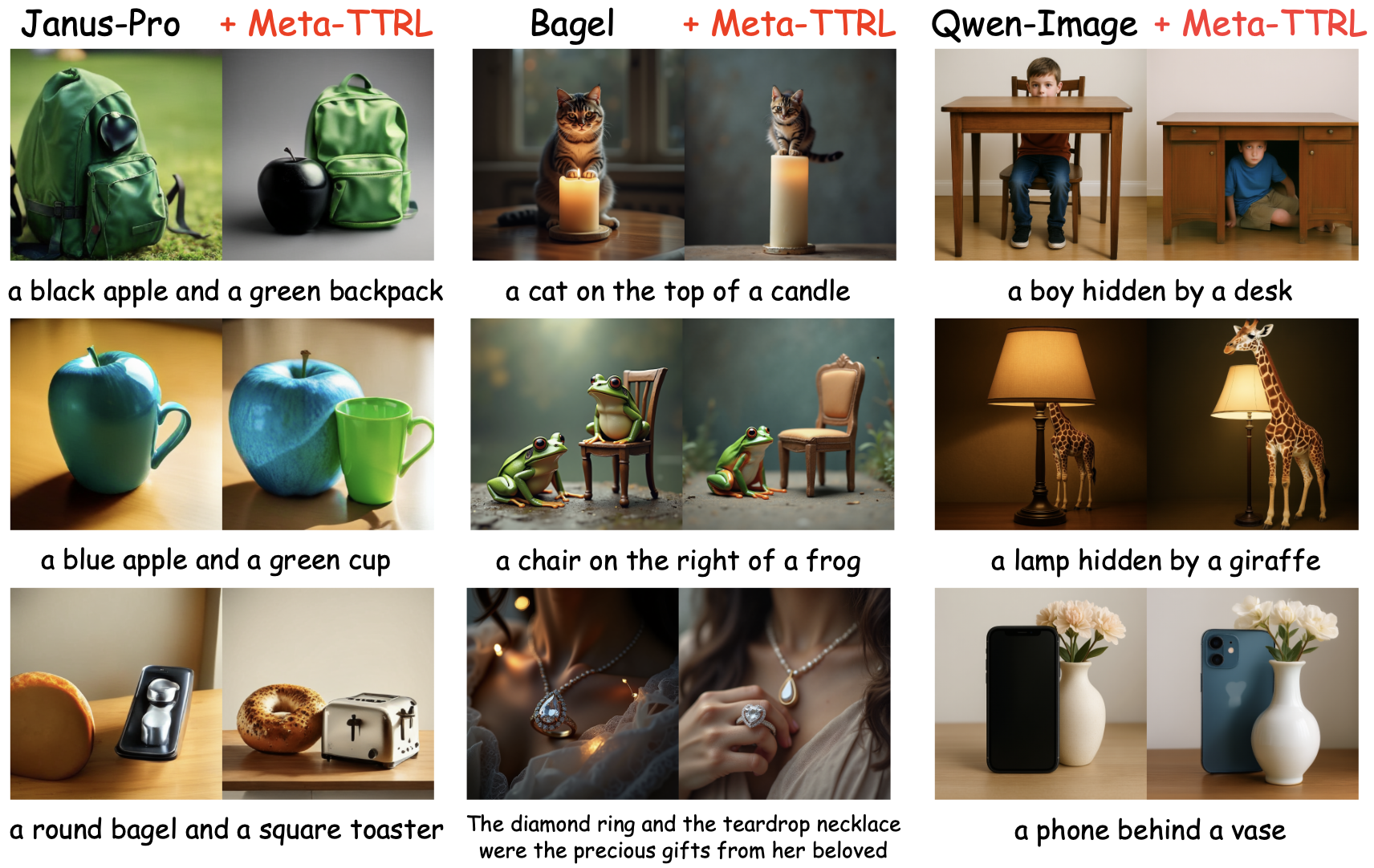} 
    \caption{Qualitative case studies of \MetaTTRL}  
    \vspace{-10pt} 
    \label{fig:case_study} 
\end{figure}
\textbf{(1)} \textbf{Meta-TTRL} consistently improves performance across all evaluated models, including Qwen-Image, BAGEL, and Janus-Pro-7B. Notably, improvements are observed not only for weaker baselines but also for costly post-trained UMMs like Qwen-Image and BAGEL. Specifically, \textbf{Meta-TTRL} achieves average gains of 2.01, 4.30, and 5.41 for Qwen-Image, BAGEL, and Janus-Pro-7B, respectively, on the TIIF-Bench. In addition, \textbf{Meta-TTRL} results in significant total gains of 10.34, 22.55, and 109.25 on T2I-CompBench++ across eight compositional subdimensions for the three evaluated models. 

\textbf{(2)} \textbf{Meta-TTRL} demonstrates pronounced improvements in attributes that require compositional reasoning, such as shape, texture, spatial reasoning, and numeracy. For instance, Qwen-Image shows improvements of 4.28\% on shape and 5.17\% on 2D spatial. BAGEL achieves gains of 6.90\%, 11.90\%, and 15.64\% on shape, 2D spatial, and 3D spatial attributes, respectively. These improvements indicate that meta-knowledge provides useful and effective guidance during training, enabling UMMs to better capture attribute dependencies and relational structures in T2I generation.

\textbf{(3)} We observe that \textbf{Meta-TTRL} leads to more dramatic relative improvements for models with weaker compositional reasoning at baseline. For example, Janus-Pro-7B, which lags behind Qwen-Image and BAGEL on T2ICompBench++, benefits from significant improvements such as 52.50\% on color, 53.12\% on shape, 67.17\% on texture, and 106.36\% on 2D spatial. In comparison, stronger models like Qwen-Image and BAGEL show moderate but consistently positive improvements. We attribute this to the monitoring signals in \textbf{Meta-TTRL}, which are particularly beneficial for models with weaker generative performance at baseline, as this metacognitive mechanism leverages the model’s meta-knowledge to guide object-level generation.

\textbf{(4)} Beyond benchmark-specific improvements, \textbf{Meta-TTRL} generalizes well to out-of-distribution settings. We first evaluate its generalization on the GenEval benchmark~\cite{ghosh2023geneval}. As shown in Table~\ref{tab:geneval_OOD}, \textbf{Meta-TTRL} consistently improves performance across all three evaluated models. Further, we evaluate cross-benchmark generalization using Janus-Pro-7B: we apply \textbf{Meta-TTRL} on each T2I benchmark and evaluating T2I performance on the others (Table~\ref{tab:OOD_Evaluation}). Despite substantial differences in prompt distributions, evaluation metrics, and judging protocols, \textbf{Meta-TTRL} maintains consistent improvements across all settings, indicating that \textbf{Meta-TTRL} does not overfit to specific benchmarks and acquires generalizable gains in image generation quality.

\begin{table}[h]
\caption{Performance on GenEval for all evaluated models}
\label{tab:geneval_OOD}
\centering
\begin{small} 
\begin{sc}
\begin{tabular}{lccc}
\toprule
Model & Train on & Vanilla & \MetaTTRL \\
\midrule
Janus-Pro-7B & T2I++  & \multirow{3}{*}{77.80} & 81.14 {\footnotesize\textbf{\textcolor{ForestGreen}{+3.34}}} \\
             & TIIF   &                        & 79.40 {\footnotesize\textbf{\textcolor{ForestGreen}{+1.60}}} \\
             & DPG    &                        & 78.45 {\footnotesize\textbf{\textcolor{ForestGreen}{+0.65}}} \\
\midrule
Bagel        & T2I++  & \multirow{3}{*}{79.24} & 83.37 {\footnotesize\textbf{\textcolor{ForestGreen}{+4.13}}} \\
             & TIIF   &                        & 82.67 {\footnotesize\textbf{\textcolor{ForestGreen}{+3.43}}} \\
             & DPG    &                        & 82.67 {\footnotesize\textbf{\textcolor{ForestGreen}{+3.43}}} \\
\midrule
Qwem-Image   & T2I++  & \multirow{3}{*}{87.00} & 88.52 {\footnotesize\textbf{\textcolor{ForestGreen}{+1.52}}} \\
             & TIIF   &                        & 89.72 {\footnotesize\textbf{\textcolor{ForestGreen}{+2.72}}} \\
             & DPG    &                        & 88.43 {\footnotesize\textbf{\textcolor{ForestGreen}{+1.43}}} \\
\bottomrule
\vspace{-18pt}
\end{tabular}
\end{sc}
\end{small}  
\end{table} 
\begin{table}[h]
  \caption{Cross-benchmark generalization on Janus-Pro-7B.}
  \label{tab:OOD_Evaluation}
  \centering
  \begin{small}
    \begin{sc}
      \begin{tabular}{lccr}
        \toprule
        Train on & Eval on & Vanilla & \MetaTTRL \\
        \midrule
        \multirow{2}{*}{T2I++}
          & TIIF    & 65.68 & 68.71 {\small\textcolor{ForestGreen}{\textbf{+3.03}}} \\
          & DPG     & 83.84 & 84.63 {\small\textcolor{ForestGreen}{\textbf{+0.79}}} \\ 
        \midrule
        \multirow{2}{*}{TIIF}
          & T2I++   & 35.25 & 40.44 {\small\textcolor{ForestGreen}{\textbf{+5.19}}} \\
          & DPG     & 83.84 & 84.32 {\small\textcolor{ForestGreen}{\textbf{+0.48}}} \\ 
        \midrule
        \multirow{2}{*}{DPG}
          & T2I++   & 35.25 & 38.93 {\small\textcolor{ForestGreen}{\textbf{+3.68}}} \\
          & TIIF    & 65.68 & 66.71 {\small\textcolor{ForestGreen}{\textbf{+1.03}}} \\
        \bottomrule
        \vspace{-18pt}
      \end{tabular}
    \end{sc}
  \end{small}
\end{table}

\begin{figure*}[t]  
    \centering
    \begin{subfigure}{0.245\textwidth}
        \centering
        \includegraphics[width=\textwidth]{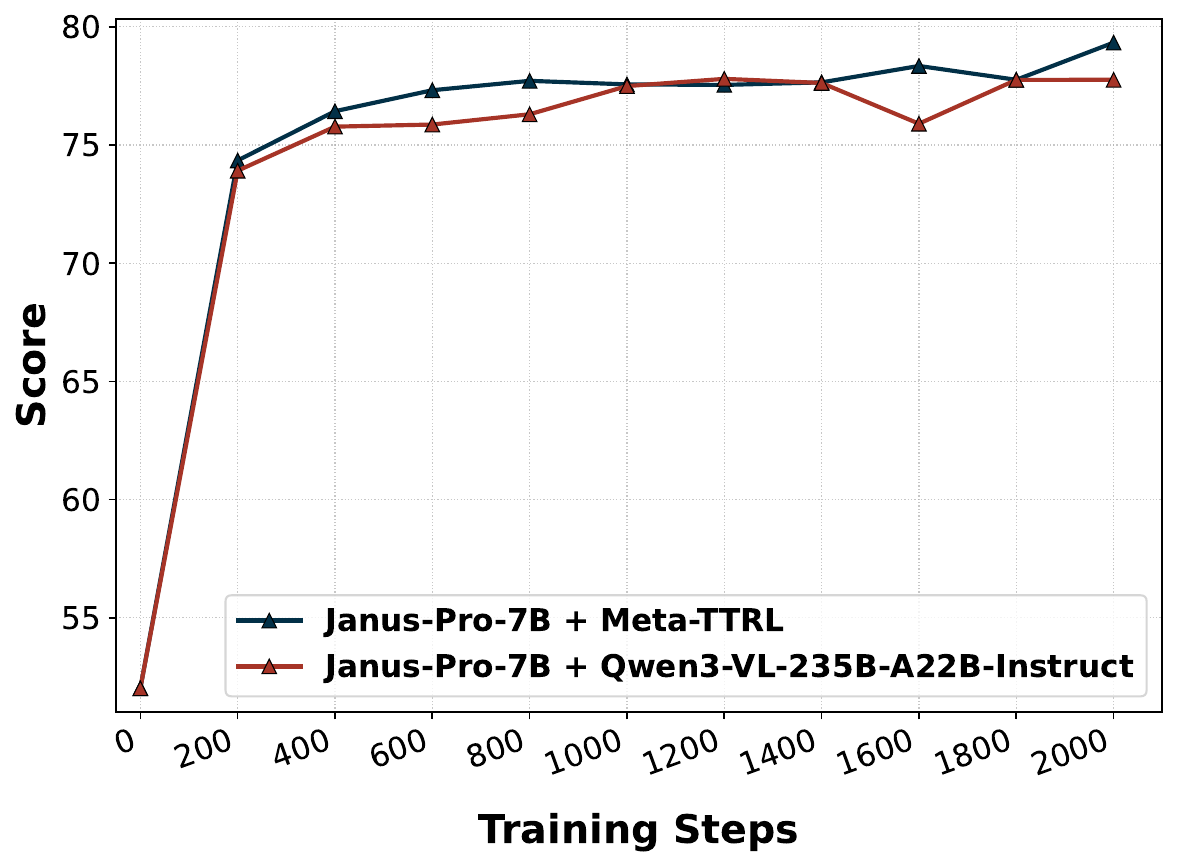}
        \caption{Color}
        \label{subfig:color}
    \end{subfigure}
    \hfill
    \begin{subfigure}{0.245\textwidth}
        \centering
        \includegraphics[width=\textwidth]{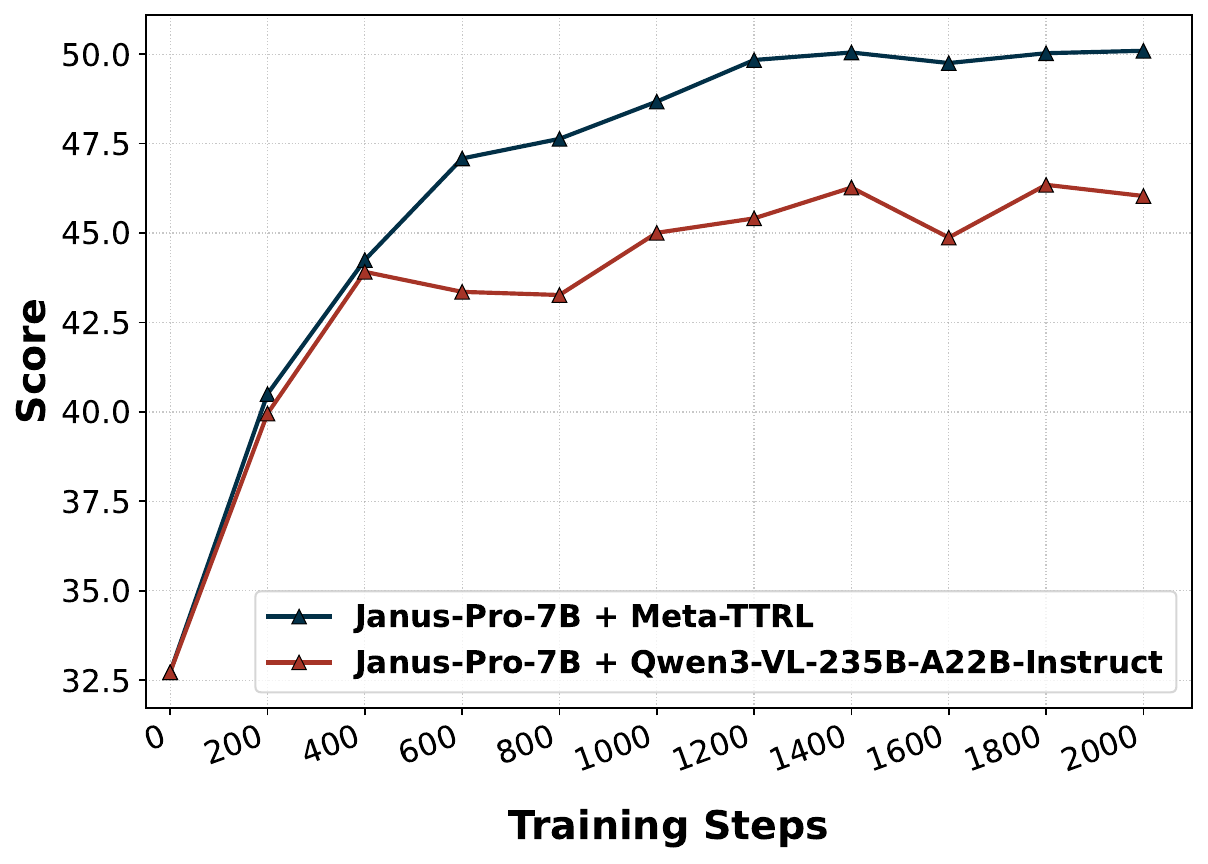}
        \caption{Shape}
        \label{subfig:shape}
    \end{subfigure}
    \hfill
    \begin{subfigure}{0.245\textwidth}
        \centering
        \includegraphics[width=\textwidth]{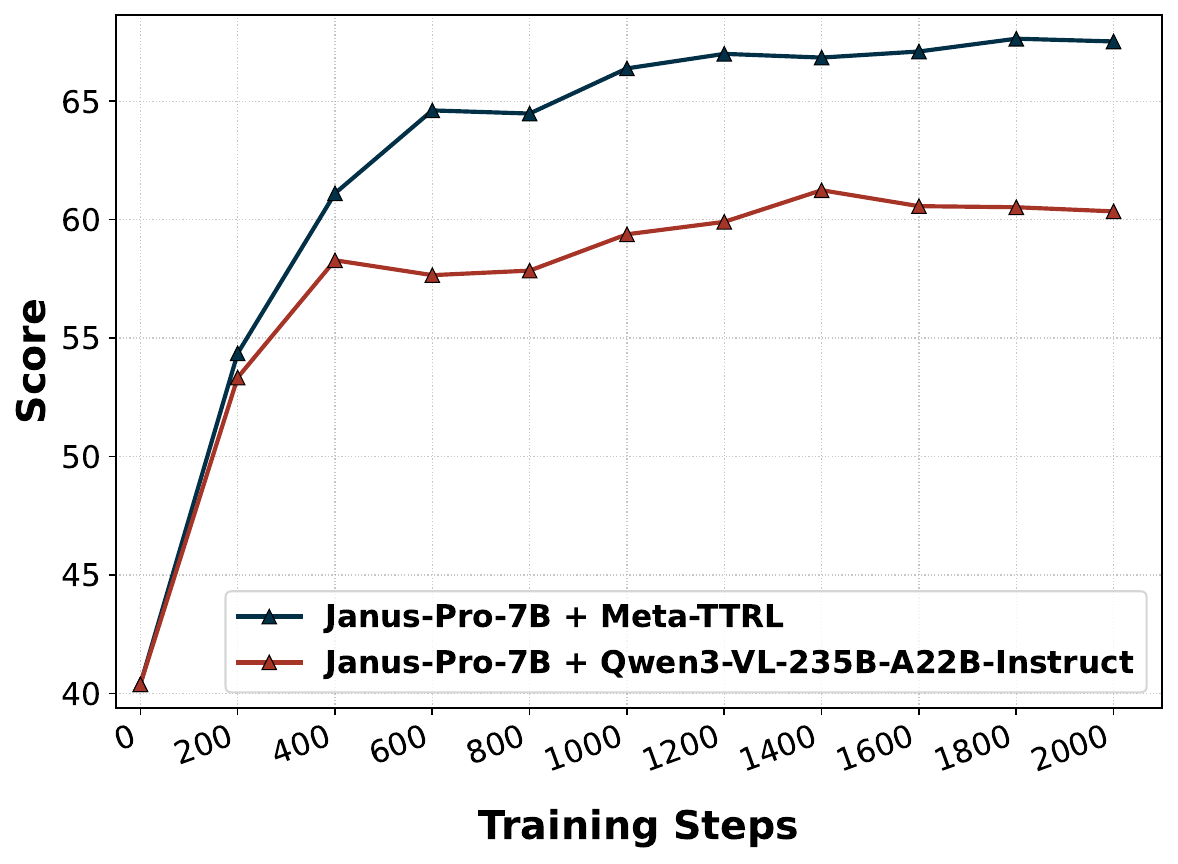}
        \caption{Texture}
        \label{subfig:texture}
    \end{subfigure}
    \hfill
    \begin{subfigure}{0.245\textwidth}
        \centering
        \includegraphics[width=\textwidth]{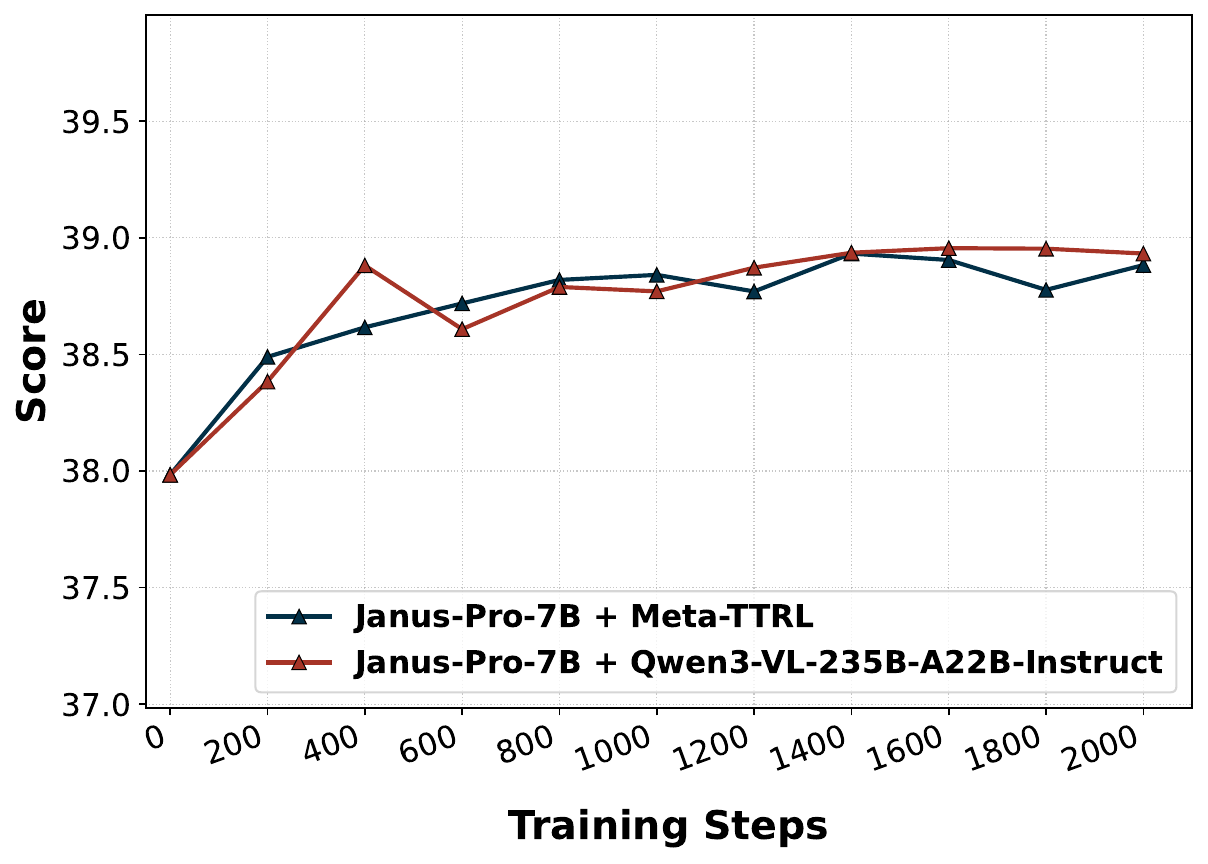}
        \caption{Complex}
        \label{subfig:complex}
    \end{subfigure}

    \vspace{0.2cm}

    \begin{subfigure}{0.245\textwidth}
        \centering
        \includegraphics[width=\textwidth]{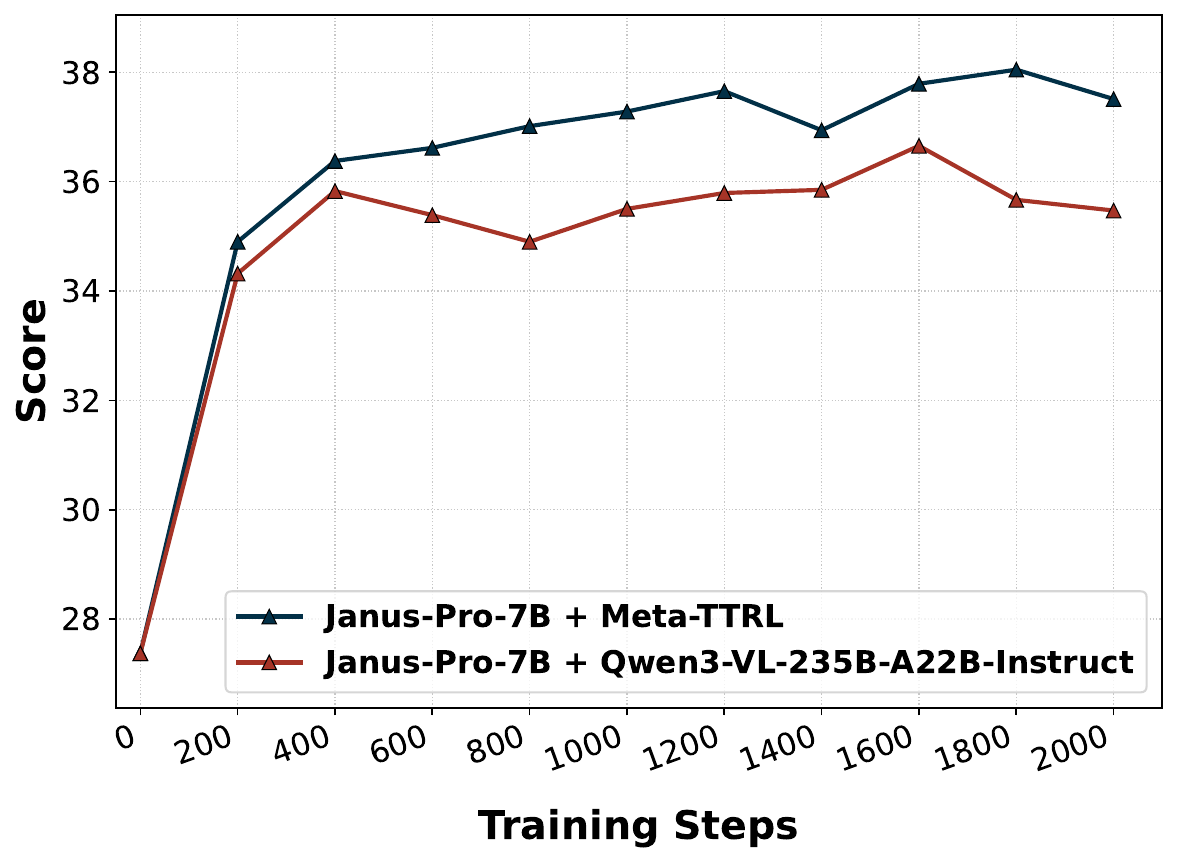}
        \caption{3D Spatial}
        \label{subfig:3dspatial}
    \end{subfigure}
    \hfill
    \begin{subfigure}{0.245\textwidth}
        \centering
        \includegraphics[width=\textwidth]{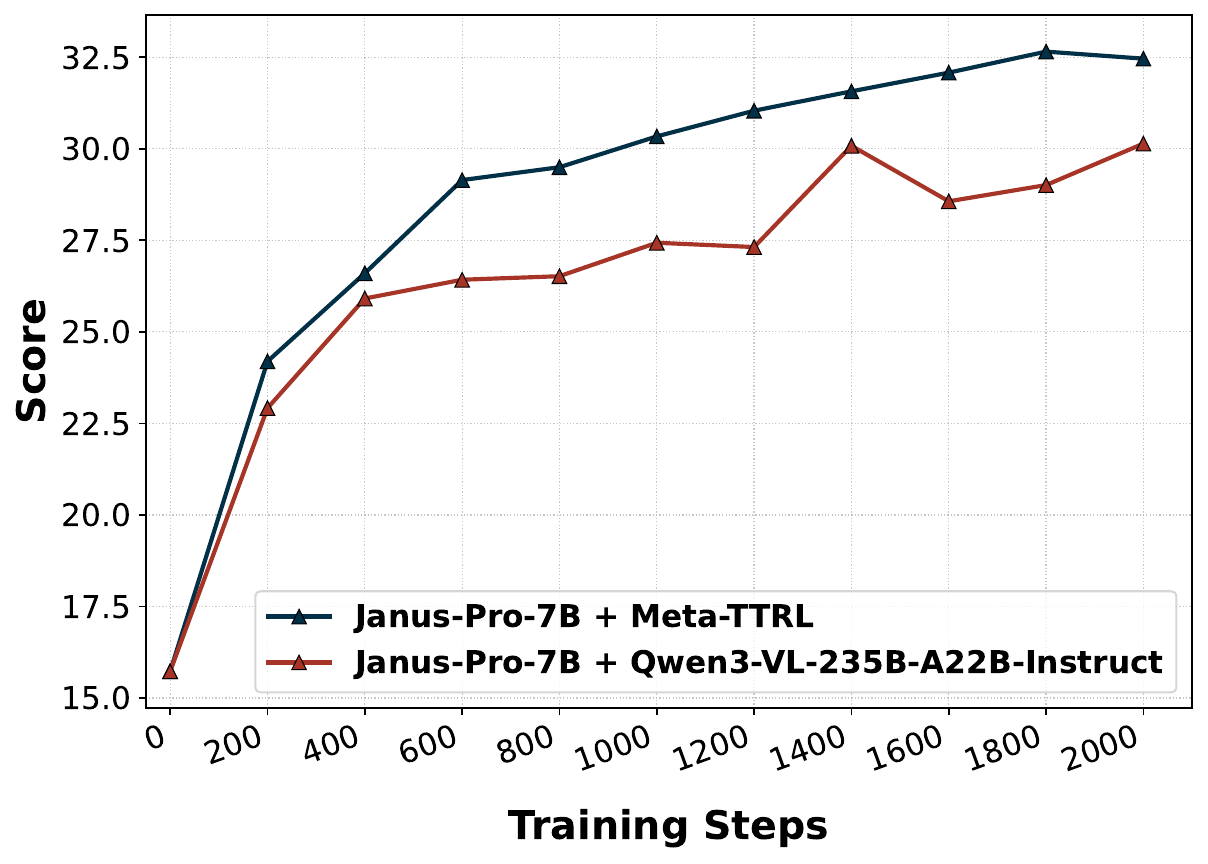}
        \caption{2D Spatial}
        \label{subfig:spatial}
    \end{subfigure}
    \hfill
    \begin{subfigure}{0.245\textwidth}
        \centering
        \includegraphics[width=\textwidth]{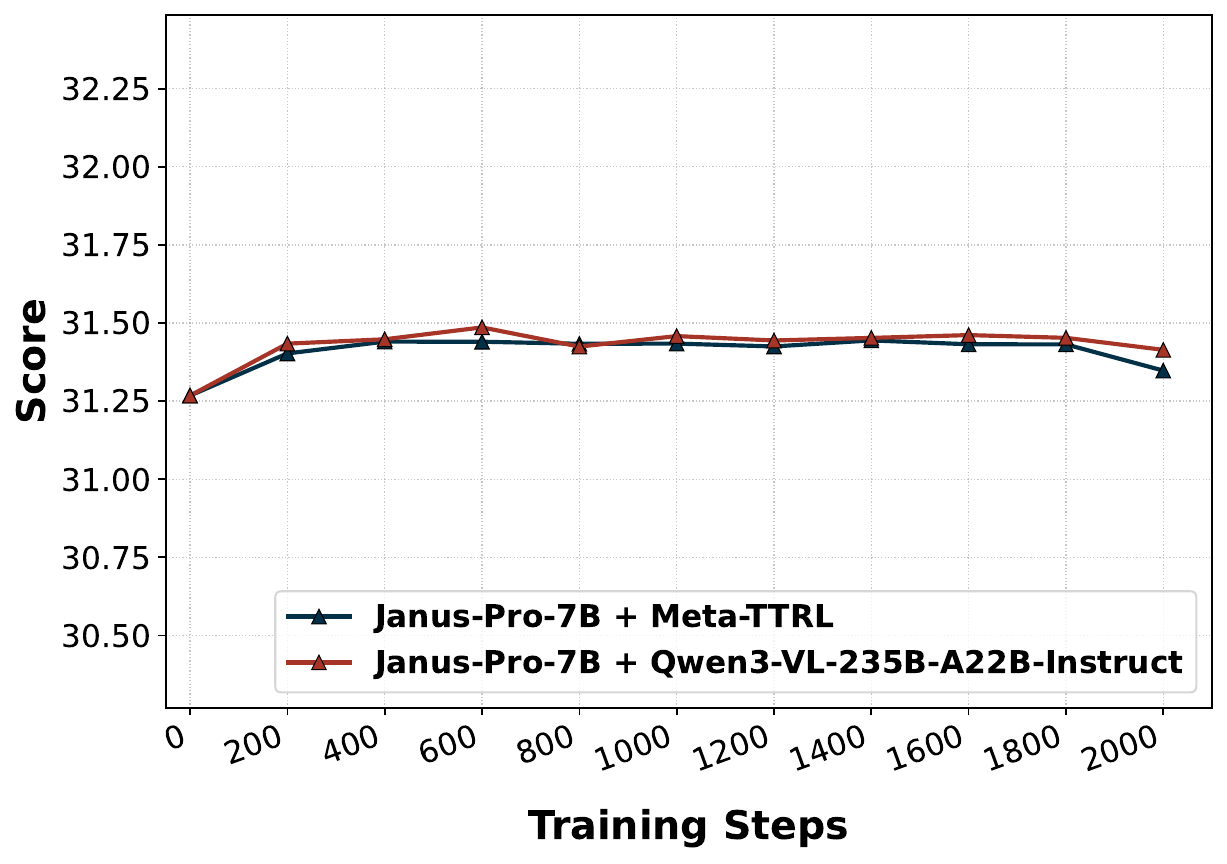}
        \caption{Non-spatial}
        \label{subfig:nonspatial}
    \end{subfigure}
    \hfill
    \begin{subfigure}{0.245\textwidth}
        \centering
        \includegraphics[width=\textwidth]{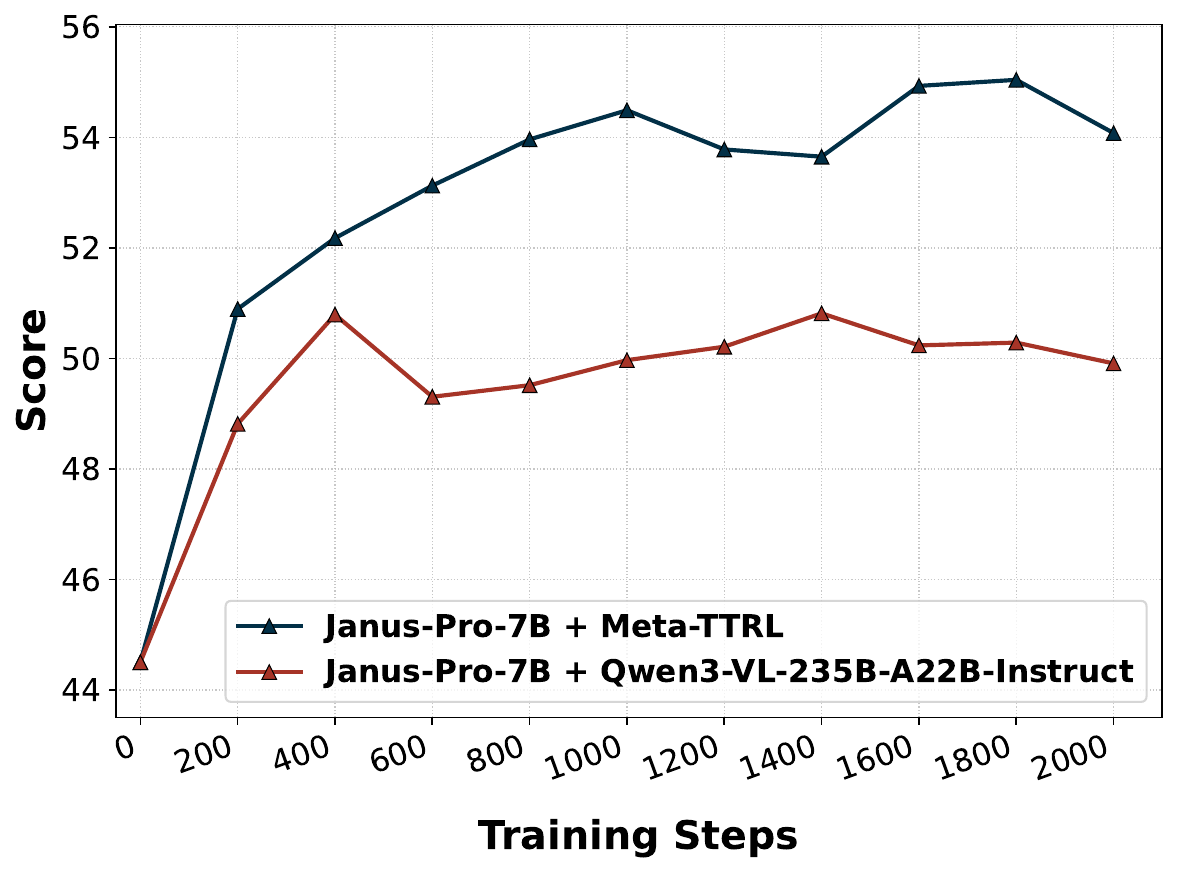}
        \caption{Numeracy}
        \label{subfig:numeracy}
    \end{subfigure}
       \caption{Performance comparison between \textbf{Meta-TTRL (blue)} and \textbf{E-TTRL (red)} on T2I-CompBench++ across eight subdimensions.}
    \label{fig:qwen3_performance}
\end{figure*}

\section{Analysis}
In this section, we investigate the \emph{potential of TTRL for improving T2I generation} in UMMs. T2I generation is inherently ill-posed: a single textual prompt may correspond to many plausible images, making theoretical characterization of performance limits difficult. Instead, we study the potential of TTRL empirically through a series of controlled analyses. Specifically, we analyze three aspects through comparisons with \textbf{Meta-TTRL}:
\textbf{(1) External Introspector (E-TTRL)}: the introspector is replaced with a large-scale multimodal model, Qwen3-VL-235B-A22B-Instruct~\cite{Qwen3-VL}, to study the effect of introspector capability on TTRL; \textbf{(2) RL Leakage}: reinforcement learning is performed directly on test prompts using UnifiedReward~\cite{unifiedreward}, a widely used reward model for image generation, providing an approximate empirical upper bound for TTRL;
\textbf{(3) Alternative Monitoring Signals}: rubric construction is retained with the introspector while rubric evaluation is performed by the visual question answering (VQA) model, GIT~\cite{wang2022git}, to examine whether alternative monitoring mechanisms can support effective TTRL. To ensure fair comparison, we fix the number of reinforcement learning steps for each model on each benchmark across all experiments (see Appendix \textcolor{linkblue}{D} for details).

\subsection{External Introspector (E-TTRL)}
\emph{Does a stronger external introspector improve TTRL?} 
A natural hypothesis is that leveraging the stronger reasoning and multimodal understanding of an external model may provide more informative signals and thus lead to more effective TTRL. To test this hypothesis, we perform E-TTRL, where the introspector in \textbf{Meta-TTRL} is replaced with Qwen3-VL-235B-A22B-Instruct. We use Janus-Pro-7B as the base model. The monitoring–control pipeline remains unchanged, while both rubric construction and rubric evaluation are performed by the external model. We compare E-TTRL with \textbf{Meta-TTRL} on T2I-CompBench++, and Figure~\ref{fig:qwen3_performance} plots the performance of both methods across training steps on the eight subdimensions. As shown in Figure~\ref{fig:qwen3_performance}, \textbf{Meta-TTRL} consistently outperforms E-TTRL on six out of eight subdimensions. 
Notably, \textbf{Meta-TTRL} achieves substantial gains in subdimensions such as shape, texture, and numeracy, while performance on the remaining two subdimensions remains comparable without significant differences. We attribute this to the monitoring signals provided by the external introspector in E-TTRL often lie beyond the effective optimization regime of the base model. This misalignment limits its utility for gradient-based test-time optimization. In contrast, intrinsic introspection in \textbf{Meta-TTRL} aligns more effectively with the base model’s optimization regime, facilitating a more stable and efficient learning process, inducing a curriculum-like test-time learning process. These results suggest that the success of T2I TTRL depends less on the absolute strength of the evaluator and more on the capacity-matched introspective signals that align with the model's effective optimization regime. \textbf{Insight:} the effectiveness of TTRL depends less on the absolute capability of the evaluator and more on whether the monitoring signals are \emph{capacity-matched}, highlighting the critical role of \emph{metacognitive synergy} in enhancing TTRL performance.

\begin{figure*}[t]
    \centering
    \includegraphics[width=0.595\textwidth]{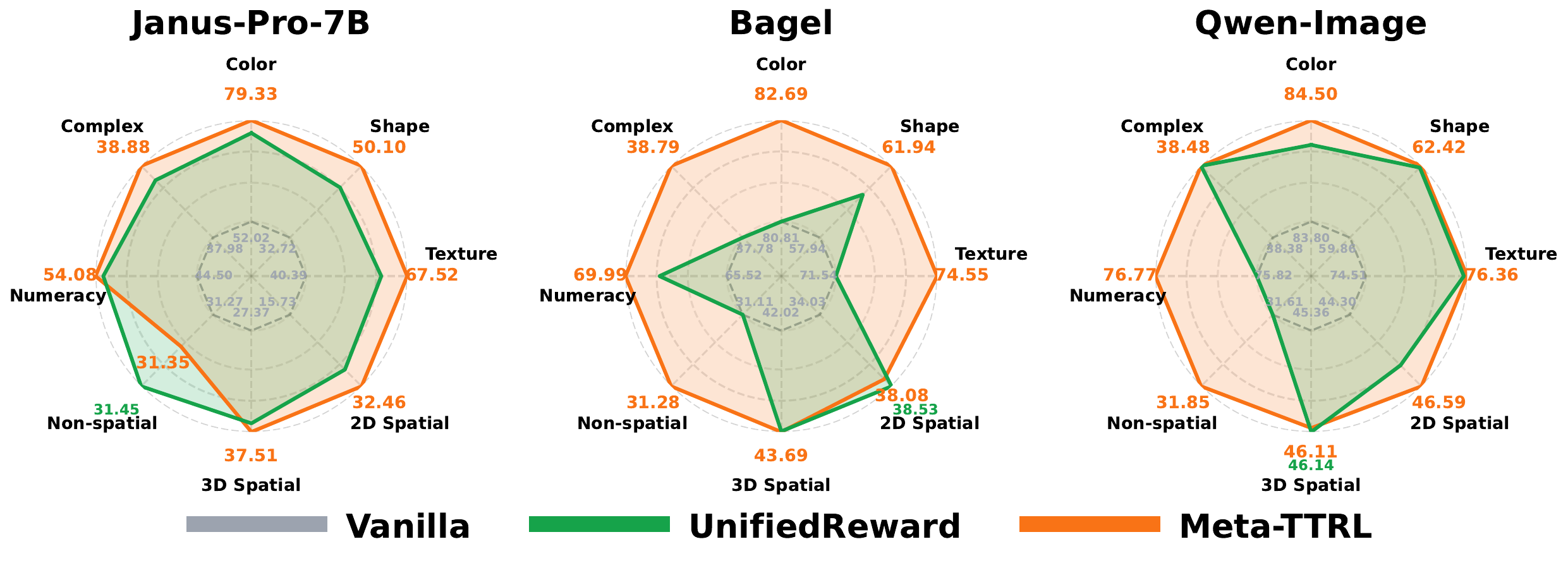}
    \includegraphics[
      trim=7 5 5 0,
      clip,
      width=0.395\textwidth
    ]{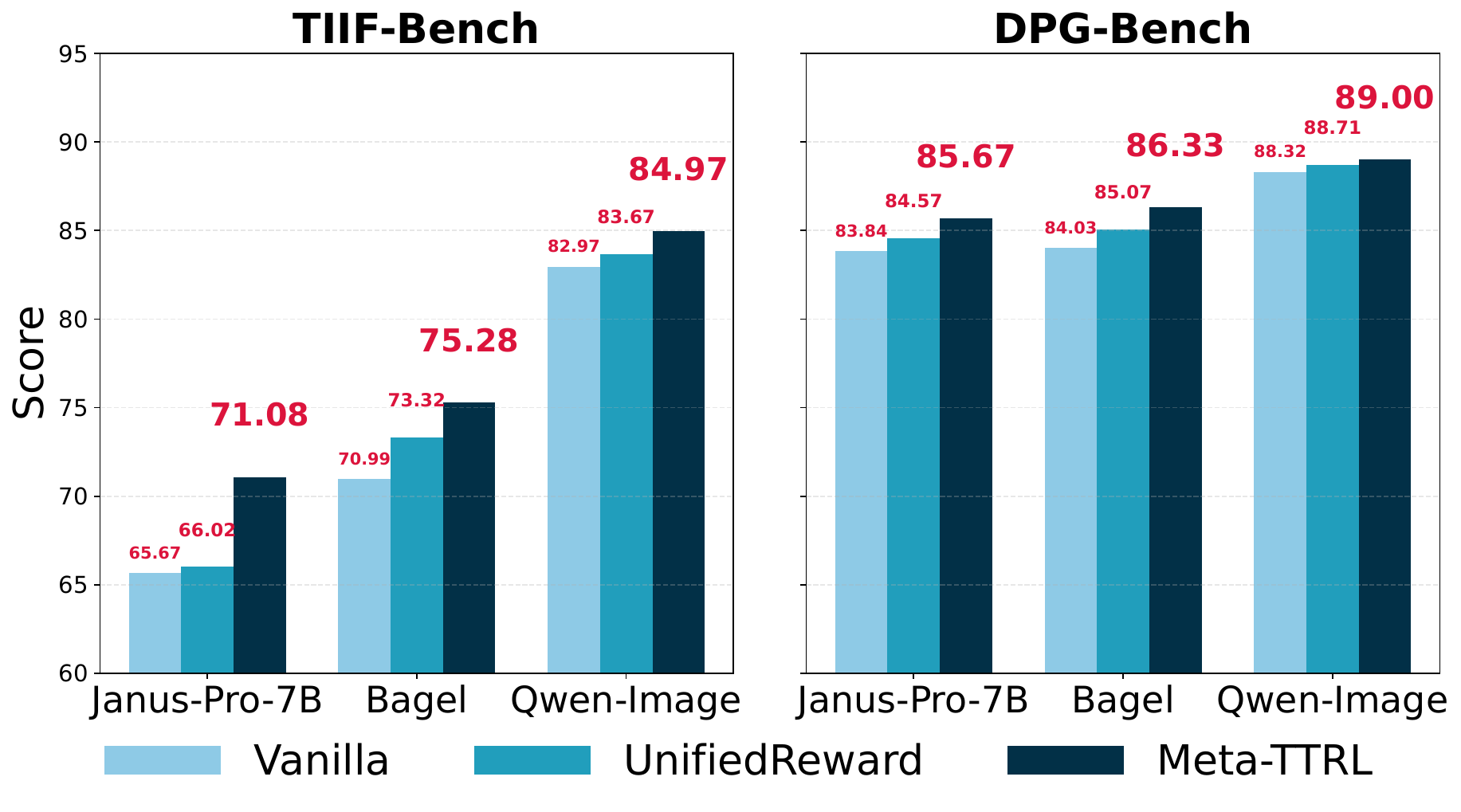}
    \caption{
    Performance comparison of Baseline, RL Leakage with UnifiedReward, and \textbf{Meta-TTRL} across three T2I benchmarks. \\
    \textbf{Left:}  Radar plot comparing performance on T2I-CompBench++.
    \textbf{Right:} Bar charts showing results on TIIF-Bench and DPG-Bench.}
    \label{fig:radar_and_bar}
\end{figure*}

\begin{table*}[t]
\caption{Comparison between \textbf{Meta-TTRL} and replacing rubric evaluation with GIT across three benchmarks.}
\label{tab:git_table}
\centering
\setlength{\tabcolsep}{3pt} 
\begin{footnotesize}
\begin{sc}
\begin{tabular}{lccccccccc}
\toprule
Train on & \multicolumn{3}{c}{Janus-Pro-7B} & \multicolumn{3}{c}{Bagel} & \multicolumn{3}{c}{Qwem-Image} \\
\cmidrule(lr){2-4} \cmidrule(lr){5-7} \cmidrule(lr){8-10}
         & Vanilla & GIT & \textbf{Meta-TTRL} & Vanilla & GIT & \textbf{Meta-TTRL} & Vanilla & GIT & \textbf{Meta-TTRL} \\
\midrule
T2I++  
& 35.25 & 47.16 {\tiny\textbf{\textcolor{gray}{+11.91}}} & 48.90 {\footnotesize\textbf{\textcolor{ForestGreen}{+13.65}}} 
& 52.60 & 54.83  {\tiny\textbf{\textcolor{gray}{+2.23}}} & 55.13 {\footnotesize\textbf{\textcolor{ForestGreen}{+2.53}}} 
& 56.71 &57.71  {\tiny\textbf{\textcolor{gray}{+1.00}}} & 57.88 {\footnotesize\textbf{\textcolor{ForestGreen}{+1.17}}} \\
TIIF 
& 65.68 & 69.73  {\tiny\textbf{\textcolor{gray}{+4.05}}} & 71.08  {\footnotesize\textbf{\textcolor{ForestGreen}{+5.40}}} 
& 71.00  & 72.70 {\tiny\textbf{\textcolor{gray}{+1.70}}} & 75.29  {\footnotesize\textbf{\textcolor{ForestGreen}{+4.29}}} 
& 82.97 & 84.24 {\tiny\textbf{\textcolor{gray}{+1.27}}} & 84.98 {\footnotesize\textbf{\textcolor{ForestGreen}{+2.01}}} \\
DPG 
& 83.84 & 84.78 {\tiny\textbf{\textcolor{gray}{+0.94}}} & 85.67 {\footnotesize\textbf{\textcolor{ForestGreen}{+1.83}}} 
& 84.03 &86.24 {\tiny\textbf{\textcolor{gray}{+2.21}}} & 86.33 {\footnotesize\textbf{\textcolor{ForestGreen}{+2.30}}} 
& 88.32 & 88.83 {\tiny\textbf{\textcolor{gray}{+0.51}}} & 89.00{\footnotesize\textbf{\textcolor{ForestGreen}{+0.68}}} \\
\bottomrule
\vspace{-16pt}
\end{tabular}
\end{sc}
\end{footnotesize}
\end{table*} 

\subsection{RL Leakage}
\emph{How much improvement can RL leakage bring to TTRL?} 
To further explore the potential of TTRL, we consider RL leakage, a setting where reinforcement learning is directly applied to the test prompts using UnifiedReward, a widely used reward model in post-training reinforcement learning for image generation. Although this violates the standard train–test separation followed in post-training reinforcement learning, it allows us to study how much improvement RL leakage can bring to TTRL. We conduct RL Leakage experiments using Janus-Pro-7B, Bagel, and Qwen-Image as base models and evaluate on three T2I benchmarks. Figure~\ref{fig:radar_and_bar} presents the results. The left radar plots show the performance of the base model, RL Leakage with UnifiedReward, and \textbf{Meta-TTRL} across eight subdimensions on T2I-CompBench++ for Janus-Pro-7B, Bagel, and Qwen-Image. The right bar charts report results on TIIF-Bench and DPG-Bench. Overall, across all evaluated models, \textbf{Meta-TTRL} consistently outperforms RL Leakage on TIIF-Bench and DPG-Bench, and achieves superior performance on the majority of subdimensions in T2I-CompBench++. On the remaining dimensions, \textbf{Meta-TTRL} and RL Leakage exhibit comparable performance. These results suggest that the monitoring signals produced by \textbf{Meta-TTRL} are more effective for TTRL than those provided by a specially trained reward model. This advantage arises from \emph{metacognitive synergy} in \textbf{Meta-TTRL}, which enables more effective self-improvement at test time.

\subsection{Alternative Monitoring Signals}
\emph{Can alternative monitoring signals improve TTRL?}
To further investigate the role of monitoring signals in TTRL, we replace the scoring component in \textbf{Meta-TTRL} with a VQA model. VQA models can answer structured questions about visual content and have been widely used as reward models in post-training reinforcement learning for T2I generation~\cite{jiang2025t2i, park2025stablesketcher}. Specifically, we retain rubric construction with the introspector but replace rubric evaluation with the VQA model GIT. In our experiments, we adopt GIT-large-VQAv2, the large-sized version of the GIT model fine-tuned on the VQAv2 dataset~\cite{goyal2017making}. We conduct experiments using Janus-Pro-7B, Bagel, and Qwen-Image as base models and evaluate on three T2I benchmarks. Table~\ref{tab:git_table} reports the results. Overall, using GIT for rubric evaluation improves the base models but underperforms \textbf{Meta-TTRL} across the evaluated models and benchmarks. We attribute this difference to the inconsistency between rubric construction and evaluation. Rubrics are generated by the introspector but evaluated by an external VQA model, which may not fully capture the intended rubric. This mismatch introduces noise into the monitoring signals and weakens the \emph{metacognitive synergy} between monitoring and learning, eventually reducing the effectiveness of TTRL.

\subsection{Training Dynamics of Meta-TTRL}
\begin{figure}[H]
    \centering
    \includegraphics[width=0.495\textwidth]{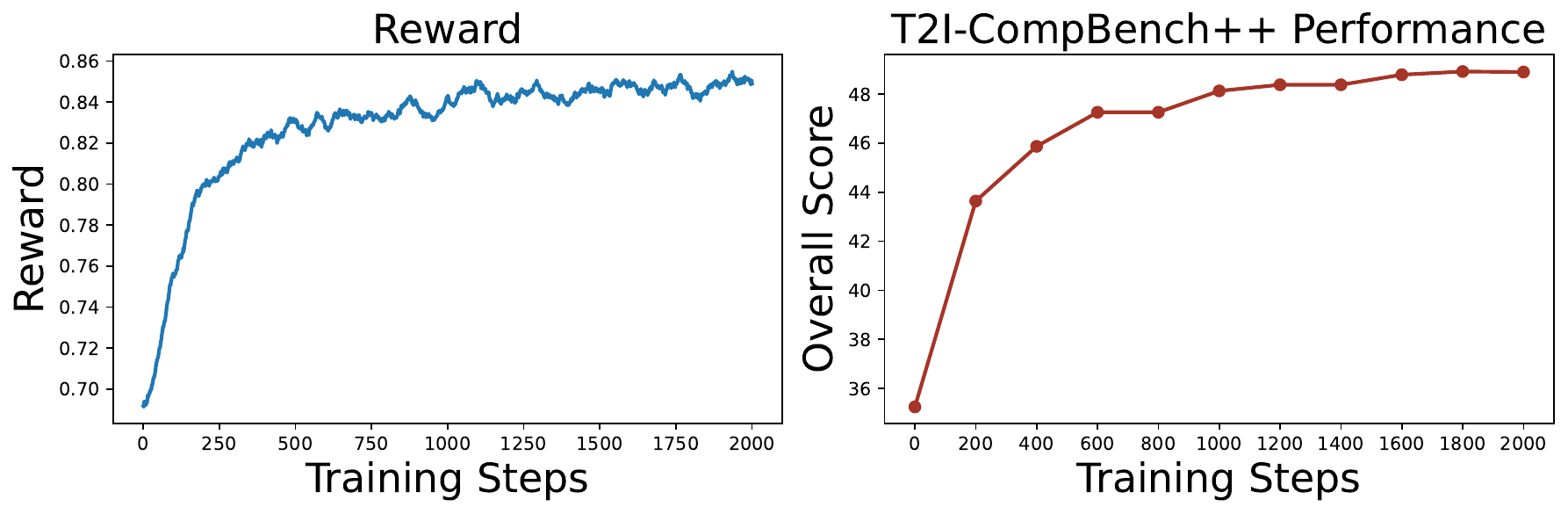}
    \caption{
   Training Dynamics of \MetaTTRL
    }
    \label{fig:reward_and_score}
     \vspace{-16pt} 
\end{figure}

To gain deeper insights into the optimization behavior of \textbf{Meta-TTRL}, we analyze the training dynamics during test-time optimization. We track the evolution of the GRPO reward and the corresponding T2I-CompBench++ average score. As depicted in Figure~\ref{fig:reward_and_score}, both the reward and the benchmark performance show steady improvement throughout the optimization process. This closely synchronized growth suggests that \textbf{Meta-TTRL} optimizes meaningful feedback signals, avoiding the reward hacking often observed in reinforcement learning scenarios. The consistent increase in both reward and performance highlights \textbf{Meta-TTRL}'s capability to achieve sustained \emph{self-improvement} during test-time optimization.

\section{Conclusion and Limitations}
\paragraph{Conclusion} In this paper, we introduce \textbf{Meta-TTRL}, a TTRL framework for T2I generation in UMMs. \textbf{Meta-TTRL} enables self-improvement by leveraging meta-knowledge, model-intrinsic monitoring signals, eliminating the need for external reward models. Our experiments demonstrate that \textbf{Meta-TTRL} consistently improves compositional reasoning and overall generation performance across multiple UMMs. The results underscore that effective test-time optimization can be achieved through capacity-matched signals, emphasizing the importance of metacognitive synergy in T2I TTRL.

\paragraph{Limitations} \textbf{Meta-TTRL} requires access to the model’s parameters, which prevents application to closed-source models.

\bibliography{reference_paper}

@article{deng2025emerging,
  title={Emerging properties in unified multimodal pretraining},
  author={Deng, Chaorui and Zhu, Deyao and Li, Kunchang and Gou, Chenhui and Li, Feng and Wang, Zeyu and Zhong, Shu and Yu, Weihao and Nie, Xiaonan and Song, Ziang and others},
  journal={arXiv preprint arXiv:2505.14683},
  year={2025}
}

@misc{wu2025qwenimagetechnicalreport,
      title={Qwen-Image Technical Report}, 
      author={Chenfei Wu and Jiahao Li and Jingren Zhou and Junyang Lin and Kaiyuan Gao and Kun Yan and Sheng-ming Yin and Shuai Bai and Xiao Xu and Yilei Chen and Yuxiang Chen and Zecheng Tang and Zekai Zhang and Zhengyi Wang and An Yang and Bowen Yu and Chen Cheng and Dayiheng Liu and Deqing Li and Hang Zhang and Hao Meng and Hu Wei and Jingyuan Ni and Kai Chen and Kuan Cao and Liang Peng and Lin Qu and Minggang Wu and Peng Wang and Shuting Yu and Tingkun Wen and Wensen Feng and Xiaoxiao Xu and Yi Wang and Yichang Zhang and Yongqiang Zhu and Yujia Wu and Yuxuan Cai and Zenan Liu},
      year={2025},
      eprint={2508.02324},
      archivePrefix={arXiv},
      primaryClass={cs.CV},
      url={https://arxiv.org/abs/2508.02324}, 
}

@misc{chen2025januspro,
      title={Janus-Pro: Unified Multimodal Understanding and Generation with Data and Model Scaling}, 
      author={Xiaokang Chen and Zhiyu Wu and Xingchao Liu and Zizheng Pan and Wen Liu and Zhenda Xie and Xingkai Yu and Chong Ruan},
      year={2025},
}

@article{unifiedreward,
  title={Unified reward model for multimodal understanding and generation},
  author={Wang, Yibin and Zang, Yuhang and Li, Hao and Jin, Cheng and Wang, Jiaqi},
  journal={arXiv preprint arXiv:2503.05236},
  year={2025}
}

@article{Qwen3-VL,
      title={Qwen3-VL Technical Report}, 
      author={Shuai Bai and Yuxuan Cai and Ruizhe Chen and Keqin Chen and Xionghui Chen and Zesen Cheng and Lianghao Deng and Wei Ding and Chang Gao and Chunjiang Ge and Wenbin Ge and Zhifang Guo and Qidong Huang and Jie Huang and Fei Huang and Binyuan Hui and Shutong Jiang and Zhaohai Li and Mingsheng Li and Mei Li and Kaixin Li and Zicheng Lin and Junyang Lin and Xuejing Liu and Jiawei Liu and Chenglong Liu and Yang Liu and Dayiheng Liu and Shixuan Liu and Dunjie Lu and Ruilin Luo and Chenxu Lv and Rui Men and Lingchen Meng and Xuancheng Ren and Xingzhang Ren and Sibo Song and Yuchong Sun and Jun Tang and Jianhong Tu and Jianqiang Wan and Peng Wang and Pengfei Wang and Qiuyue Wang and Yuxuan Wang and Tianbao Xie and Yiheng Xu and Haiyang Xu and Jin Xu and Zhibo Yang and Mingkun Yang and Jianxin Yang and An Yang and Bowen Yu and Fei Zhang and Hang Zhang and Xi Zhang and Bo Zheng and Humen Zhong and Jingren Zhou and Fan Zhou and Jing Zhou and Yuanzhi Zhu and Ke Zhu},
	  journal={arXiv preprint arXiv:2511.21631},
      year={2025}
}

@article{jiang2025t2i,
  title={T2I-R1: Reinforcing Image Generation with Collaborative Semantic-level and Token-level CoT},
  author={Jiang, Dongzhi and Guo, Ziyu and Zhang, Renrui and Zong, Zhuofan and Li, Hao and Zhuo, Le and Yan, Shilin and Heng, Pheng-Ann and Li, Hongsheng},
  journal={arXiv preprint arXiv:2505.00703},
  year={2025}
}

@article{wei2025tiif,
  title={TIIF-Bench: How Does Your T2I Model Follow Your Instructions?},
  author={Wei, Xinyu and Zhang, Jinrui and Wang, Zeqing and Wei, Hongyang and Guo, Zhen and Zhang, Lei},
  journal={arXiv preprint arXiv:2506.02161},
  year={2025}
}

@article{huang2025t2i,
  title={T2i-compbench++: An enhanced and comprehensive benchmark for compositional text-to-image generation},
  author={Huang, Kaiyi and Duan, Chengqi and Sun, Kaiyue and Xie, Enze and Li, Zhenguo and Liu, Xihui},
  journal={IEEE Transactions on Pattern Analysis and Machine Intelligence},
  year={2025},
  publisher={IEEE}
}

@misc{hu2024ella,
      title={ELLA: Equip Diffusion Models with LLM for Enhanced Semantic Alignment}, 
      author={Xiwei Hu and Rui Wang and Yixiao Fang and Bin Fu and Pei Cheng and Gang Yu},
      year={2024},
      eprint={2403.05135},
      archivePrefix={arXiv},
      primaryClass={cs.CV}
}

@article{ghosh2023geneval,
  title={Geneval: An object-focused framework for evaluating text-to-image alignment},
  author={Ghosh, Dhruba and Hajishirzi, Hannaneh and Schmidt, Ludwig},
  journal={Advances in Neural Information Processing Systems},
  volume={36},
  pages={52132--52152},
  year={2023}
}

@article{wang2022git,
  title={Git: A generative image-to-text transformer for vision and language},
  author={Wang, Jianfeng and Yang, Zhengyuan and Hu, Xiaowei and Li, Linjie and Lin, Kevin and Gan, Zhe and Liu, Zicheng and Liu, Ce and Wang, Lijuan},
  journal={arXiv preprint arXiv:2205.14100},
  year={2022}
}

@article{cui2025emu3,
  title={Emu3. 5: Native multimodal models are world learners},
  author={Cui, Yufeng and Chen, Honghao and Deng, Haoge and Huang, Xu and Li, Xinghang and Liu, Jirong and Liu, Yang and Luo, Zhuoyan and Wang, Jinsheng and Wang, Wenxuan and others},
  journal={arXiv preprint arXiv:2510.26583},
  year={2025}
}

@article{achiam2023gpt,
  title={Gpt-4 technical report},
  author={Achiam, Josh and Adler, Steven and Agarwal, Sandhini and Ahmad, Lama and Akkaya, Ilge and Aleman, Florencia Leoni and Almeida, Diogo and Altenschmidt, Janko and Altman, Sam and Anadkat, Shyamal and others},
  journal={arXiv preprint arXiv:2303.08774},
  year={2023}
}

@article{zuo2025ttrl,
  title={Ttrl: Test-time reinforcement learning},
  author={Zuo, Yuxin and Zhang, Kaiyan and Sheng, Li and Qu, Shang and Cui, Ganqu and Zhu, Xuekai and Li, Haozhan and Zhang, Yuchen and Long, Xinwei and Hua, Ermo and others},
  journal={arXiv preprint arXiv:2504.16084},
  year={2025}
}

@article{tian2025unigen,
  title={UniGen: Enhanced Training \& Test-Time Strategies for Unified Multimodal Understanding and Generation},
  author={Tian, Rui and Gao, Mingfei and Xu, Mingze and Hu, Jiaming and Lu, Jiasen and Wu, Zuxuan and Yang, Yinfei and Dehghan, Afshin},
  journal={arXiv preprint arXiv:2505.14682},
  year={2025}
}

@article{singh2025ttrv,
  title={TTRV: Test-Time Reinforcement Learning for Vision Language Models},
  author={Singh, Akshit and Marjit, Shyam and Lin, Wei and Gavrikov, Paul and Yeung-Levy, Serena and Kuehne, Hilde and Feris, Rogerio and Doveh, Sivan and Glass, James and Mirza, M Jehanzeb},
  journal={arXiv preprint arXiv:2510.06783},
  year={2025}
}

@article{guo2025can,
  title={Can We Generate Images with CoT? Let's Verify and Reinforce Image Generation Step by Step},
  author={Guo, Ziyu and Zhang, Renrui and Tong, Chengzhuo and Zhao, Zhizheng and Huang, Rui and Zhang, Haoquan and Zhang, Manyuan and Liu, Jiaming and Zhang, Shanghang and Gao, Peng and others},
  journal={arXiv preprint arXiv:2501.13926},
  year={2025}
}

@article{wang2025imagent,
  title={ImAgent: A Unified Multimodal Agent Framework for Test-Time Scalable Image Generation},
  author={Wang, Kaishen and Chen, Ruibo and Zheng, Tong and Huang, Heng},
  journal={arXiv preprint arXiv:2511.11483},
  year={2025}
}

@incollection{nelson1990metamemory,
  title={Metamemory: A theoretical framework and new findings},
  author={Nelson, Thomas O},
  booktitle={Psychology of learning and motivation},
  volume={26},
  pages={125--173},
  year={1990},
  publisher={Elsevier}
}

@article{xie2024show,
  title={Show-o: One single transformer to unify multimodal understanding and generation},
  author={Xie, Jinheng and Mao, Weijia and Bai, Zechen and Zhang, David Junhao and Wang, Weihao and Lin, Kevin Qinghong and Gu, Yuchao and Chen, Zhijie and Yang, Zhenheng and Shou, Mike Zheng},
  journal={arXiv preprint arXiv:2408.12528},
  year={2024}
}

@article{wang2024emu3,
  title={Emu3: Next-token prediction is all you need},
  author={Wang, Xinlong and Zhang, Xiaosong and Luo, Zhengxiong and Sun, Quan and Cui, Yufeng and Wang, Jinsheng and Zhang, Fan and Wang, Yueze and Li, Zhen and Yu, Qiying and others},
  journal={arXiv preprint arXiv:2409.18869},
  year={2024}
}

@article{liangyu2026unit,
  title={UniT: Unified Multimodal Chain-of-Thought Test-time Scaling},
  author={Liangyu Chen, Leon and Ma, Haoyu and Fan, Zhipeng and Huang, Ziqi and Sinha, Animesh and Dai, Xiaoliang and Wang, Jialiang and He, Zecheng and Yang, Jianwei and Li, Chunyuan and others},
  journal={arXiv e-prints},
  pages={arXiv--2602},
  year={2026}
}

@article{team2024chameleon,
  title={Chameleon: Mixed-modal early-fusion foundation models, 2024},
  author={Team, Chameleon},
  journal={URL https://arxiv. org/abs/2405.09818},
  volume={9},
  number={8},
  year={2024}
}

@article{dong2023dreamllm,
  title={Dreamllm: Synergistic multimodal comprehension and creation},
  author={Dong, Runpei and Han, Chunrui and Peng, Yuang and Qi, Zekun and Ge, Zheng and Yang, Jinrong and Zhao, Liang and Sun, Jianjian and Zhou, Hongyu and Wei, Haoran and others},
  journal={arXiv preprint arXiv:2309.11499},
  year={2023}
}

@inproceedings{yu2024mmoe,
  title={Mmoe: Enhancing multimodal models with mixtures of multimodal interaction experts},
  author={Yu, Haofei and Qi, Zhengyang and Jang, Lawrence Keunho and Salakhutdinov, Russ and Morency, Louis-Philippe and Liang, Paul Pu},
  booktitle={Proceedings of the 2024 Conference on Empirical Methods in Natural Language Processing},
  pages={10006--10030},
  year={2024}
}

@article{park2025stablesketcher,
  title={StableSketcher: Enhancing Diffusion Model for Pixel-based Sketch Generation via Visual Question Answering Feedback},
  author={Park, Jiho and Choi, Sieun and Seo, Jaeyoon and Kim, Jihie},
  journal={arXiv preprint arXiv:2510.20093},
  year={2025}
}

@inproceedings{goyal2017making,
  title={Making the v in vqa matter: Elevating the role of image understanding in visual question answering},
  author={Goyal, Yash and Khot, Tejas and Summers-Stay, Douglas and Batra, Dhruv and Parikh, Devi},
  booktitle={Proceedings of the IEEE conference on computer vision and pattern recognition},
  pages={6904--6913},
  year={2017}
}

@article{brown2020language,
  title={Language models are few-shot learners},
  author={Brown, Tom and Mann, Benjamin and Ryder, Nick and Subbiah, Melanie and Kaplan, Jared D and Dhariwal, Prafulla and Neelakantan, Arvind and Shyam, Pranav and Sastry, Girish and Askell, Amanda and others},
  journal={Advances in neural information processing systems},
  volume={33},
  pages={1877--1901},
  year={2020}
}

@inproceedings{ba2025enhancing,
  title={Enhancing reward models for high-quality image generation: Beyond text-image alignment},
  author={Ba, Ying and Zhang, Tianyu and Bai, Yalong and Mo, Wenyi and Liang, Tao and Su, Bing and Wen, Ji-Rong},
  booktitle={Proceedings of the IEEE/CVF International Conference on Computer Vision},
  pages={19022--19031},
  year={2025}
}

@inproceedings{mehrabi2023resolving,
  title={Resolving ambiguities in text-to-image generative models},
  author={Mehrabi, Ninareh and Goyal, Palash and Verma, Apurv and Dhamala, Jwala and Kumar, Varun and Hu, Qian and Chang, Kai-Wei and Zemel, Richard and Galstyan, Aram and Gupta, Rahul},
  booktitle={Proceedings of the 61st Annual Meeting of the Association for Computational Linguistics (Volume 1: Long Papers)},
  pages={14367--14388},
  year={2023}
}
\bibliographystyle{icml2026}

\end{document}